\documentclass[machines,article,submit,pdftex,moreauthors]{Definitions/mdpi} 
\usepackage{algorithm}
\usepackage{algpseudocode}
\usepackage{amsmath}
\usepackage{caption}
\usepackage{multirow}
\usepackage[utf8]{inputenc}
\usepackage{siunitx}
\usepackage{lineno}
\firstpage{1} 
\pubvolume{1}
\issuenum{1}
\articlenumber{0}
\pubyear{2024}
\copyrightyear{2024}
\datereceived{ } 
\daterevised{ } % Comment out if no revised date
\dateaccepted{ } 
\datepublished{ } 
\hreflink{https://doi.org/} % If needed use \linebreak
\Title{Safe Hybrid-Action Reinforcement Learning-Based Decision and Control for Discretionary Lane Change}

\TitleCitation{Title}

\Author{Ruichen Xu$^{}$\orcidA{}, Xiao Liu, Jinming Xu$^{}$\orcidB{}, Yuan Lin $^{}$*$^{}$\orcidC{}}

\AuthorNames{Firstname Lastname, Firstname Lastname and Firstname Lastname}

\AuthorCitation{Lastname, F.; Lastname, F.; Lastname, F.}
\address[1]{Shien-Ming Wu School of Intelligent Engineering, South China University of Technology, Guangzhou 510641, China.}

\corres{Correspondence: yuanlin@scut.edu.cn}

\abstract{Autonomous lane-change, a key feature of advanced driver-assistance systems, can enhance traffic efficiency and reduce the incidence of accidents. However, safe driving of autonomous vehicles remains challenging in complex environments. How to perform safe and appropriate lane change is a popular topic of research in the field of autonomous driving. Currently, few papers consider the safety of reinforcement learning in autonomous lane-change scenarios. We introduce safe hybrid-action reinforcement learning into discretionary lane change for the first time and propose Parameterized Soft Actor-Critic with PID Lagrangian (PASAC-PIDLag) algorithm. Furthermore, we conduct a comparative analysis of the Parameterized Soft Actor-Critic (PASAC), which is an unsafe version of PASAC-PIDLag. Both algorithms are employed to train the lane-change strategy of autonomous vehicles to output discrete lane-change decision and longitudinal vehicle acceleration. Our simulation results indicate that at a traffic density of 15 vehicles per kilometer (15 veh/km), the PASAC-PIDLag algorithm exhibits superior safety with a collision rate of 0\%, outperforming the PASAC algorithm, which has a collision rate of 1\%. The outcomes of the generalization assessments reveal that at low traffic density levels, both the PASAC-PIDLag and PASAC algorithms are proficient in attaining a 0\% collision rate. Under conditions of high traffic flow density, the PASAC-PIDLag algorithm surpasses PASAC in terms of both safety and optimality.}


\begin{document}
\nolinenumbers
\section{Introduction}

In recent years, with the continuous development of technology, the research and development of Advanced Driver-Assistance Systems (ADAS) have been dedicated to improving traffic safety levels by reducing the impact of human errors. According to data from the World Health Organization (WHO), road traffic accidents cause nearly 1.3 million deaths and approximately 50 million injuries worldwide each year. Extrapolating from this trend, road traffic accidents could result in approximately 13 million deaths and 500 million injuries over the next decade \cite{r1}.

With the rapid advancement of autonomous driving technology, there has been a gradual enhancement in driving comfort, safety, and user experience. Nowadays, lane change is a challenging task that necessitates precise maneuvers to ensure it is conducted safely, comfortably, and swiftly. The lane change includes both mandatory and discretionary changes \cite{r2}: (1) Mandatory lane change refers to the motion planning of lane change in situations where it is imperative to do so. Scenarios for mandatory lane change include merging from entrance ramps and changing lanes in the presence of obstacles ahead \cite{r31}. (2) Discretionary lane change are decision made by a vehicle to change lanes when it is not demanded due to road conditions, but rather motivated by factors such as speed optimization, driving efficiency, or driver preference. Unlike mandatory lane change, which occur because of immediate necessities such as road obstructions, construction, or merging, discretionary lane change offer an additional layer of complexity to autonomous vehicle algorithms.

Current scholars have provided two research methodologies for decision making in autonomous vehicle lane change: (1) rule-based methods \cite{r3,r4,r5,r36} and (2) learning-based methods \cite{r6,r7}.

(1) Rule-based decision models use a set of predefined, hand-crafted rules to simulate the decision-making process of drivers. These rules may include adherence to traffic regulations, such as stopping at red lights and proceeding at green lights. The model is highly interpretable because the rules are clear and straightforward, making them easy to understand and maintain. However, rule-based models may lack flexibility when dealing with complex driving environments and unknown situations because hand-crafted rules may not easily adapt to such complexities and uncertainties \cite{r3,r36}. 

(2) On the other hand, learning-based decision models rely on data-driven approaches and training models on large-scale driving data to autonomously learn and adapt to different driving conditions. These methods employ deep learning techniques that utilize neural networks and machine learning algorithms to address complex driving decision problems. Although this approach excels in adapting to varied driving scenarios, it has relatively poor interpretability, and there is no guarantee of safety during the training process. 
Most current learning-based articles are dedicated to using deep reinforcement learning techniques for discretionary autonomous lane-change control of self-driving vehicles \cite{r11,r12,r13,r14,r15,r16}. 

In \cite{r11}, the author proposes a framework that integrates deep reinforcement learning with Q-masking to enhance the efficiency of autonomous vehicle lane-change. In \cite{r7}, the author enhanced the efficiency of the deep Q-learning \cite{r33} algorithm and applied it to the scenario of autonomous lane change. In \cite{r12}, the author introduces an automated lane change method based on reinforcement learning, designing a Q-function approximator with a closed-form greedy policy capable of achieving smooth and efficient driving strategies in various and unpredictable scenarios. In \cite{r13}, the author develops a deep reinforcement learning agent capable of robustly executing automated lane-change in dynamic and uncertain highway environments, demonstrating superior performance over traditional heuristic-based methods through training in diverse, uncertain, and noisy traffic scenes. In \cite{r14}, the author applies deep reinforcement learning to address the challenge of successful merging or lane change for autonomous vehicles in high-density traffic, establishing a benchmark for driving in high-density traffic conditions.

%In \cite{r16}, the author proposes a deep reinforcement learning based lane-change strategy for autonomous vehicles that does not rely on Vehicle-to-Everything (V2X) communications. In \cite{r17}, the author proposes a decision-making framework for autonomous vehicles in lane-change scenarios based on deep reinforcement learning with risk awareness. These articles \cite{r11,r12,r13,r14,r15,r16} have not addressed the assurance of safety during the lane-change process. If deep reinforcement learning is to be applied to autonomous lane-change in real vehicles, ensuring the safety of decision-making is essential.

The majority of the literature currently employs discrete reinforcement learning for implementing autonomous lane change \cite{r7,r11,r12,r13,r14,r15,r16}, where the high-level control outputs lane change decision using discrete reinforcement learning, and low-level control uses car-following models such as the Intelligent Driver Model (IDM) \cite{r28} to output vehicle acceleration. The decision making and motion planning modules, as two closely adjacent and important functional modules of autonomous vehicles, are highly interrelated in terms of functionality and ultimate performance. Therefore, the design of the decision-making process should take into account the feasibility of motion planning. Likewise, motion planning should be formulated based on the decision made \cite{r30}. Therefore, in our work, we have adopted a hybrid action space to simultaneously address discrete lane change decision and continuous longitudinal acceleration control.

To apply deep reinforcement learning to autonomous lane change in real vehicles, ensuring the safety of decision-making is essential. There is a paucity of literature considering the safety aspects of autonomous lane change. Given the absence of research using safe reinforcement learning to ensure the safety of autonomous lane change, our paper uses the PID-Lagrangian based safe reinforcement learning approach \cite{r20} to implement autonomous lane change. In \cite{r17}, the author proposes a decision-making framework for autonomous vehicles in lane-change scenarios based on deep reinforcement learning with risk awareness. In \cite{r18}, the author uses a human driving lane-change decision model combined with regret theory to improve the safety and efficiency of autonomous vehicles in mixed traffic. However, neither of these studies uses the method of safe reinforcement learning. In \cite{r19}, the author introduced a safe reinforcement learning algorithm into the field of autonomous driving, combining the Proximal Policy Optimization (PPO) algorithm \cite{r34} with a PID-Lagrangian approach to enhance the traffic compliance of motion planners for self-driving vehicles.

Safe reinforcement learning \cite{r21} is a type of reinforcement learning that incorporates the concepts of safety or risk. Specifically, safe reinforcement learning emphasizes not only pursuing long-term maximum returns during the learning and implementation phases but also adhering to established safety constraints while ensuring reasonable system performance. Compared to Constrained Policy Optimization (CPO) algorithms \cite{r22} and safety reinforcement learning algorithms based on Lyapunov functions \cite{r23}, the Lagrangian-based safe reinforcement learning algorithm is simpler. Moreover, the Lagrangian-based safe reinforcement learning algorithm performed equally well or even better in tests within the Safety Gym environment \cite{r24}, despite the oscillations and overshooting observed during the learning process which can lead to constraint violations by the agent when applied in practice. Therefore, we used a PID-based Lagrangian method \cite{r20}. From a control perspective, traditional Lagrange multiplier updates behave as integral control, whereas the PID-based approach introduces proportional and differential control to stabilize the learning process of the agent.

To the best of our knowledge, there are no existing studies that apply safe hybrid action space algorithms in the domain of autonomous lane change. To bridge this gap, we introduce a methodology that aims to augment the safety and reliability of autonomous systems. The contribution of this paper includes the introduction of a novel safe hybrid action reinforcement learning algorithm, PASAC-PIDLag, and its application to the autonomous lane-change scenarios. We conducted a comprehensive and quantitative comparison between PASAC-PIDLag and PASAC, demonstrating that PASAC-PIDLag outperforms PASAC in terms of both safety and optimality. 

Regarding the structure of the paper, Section \ref{sec:algorithms} presents the PASAC-PIDLag and PASAC algorithms, Section \ref{sec:application} discusses the application of the algorithms in lane-change scenarios, Section \ref{sec:experiments} addresses the experiments and results, and Section \ref{sec:conclusions} presents the conclusions.
%%%%%%%%%%%%%%%%%%%%%%%%%%%%%%%%%%%%%%%%%%
\section{Reinforcement Learning Preliminaries}
\label{sec:algorithms}
Reinforcement learning is a computational approach for learning from interaction. In this paradigm, an agent takes actions based on the current state of the environment at each time step. As a result, the environment transitions to another state on the next time step, and the agent receives a reward based on the action taken. Both the actions taken by the agent and the rewards provided by the environment are probabilistic. The goal of an RL algorithm is to maximize the expected discounted cumulative reward over each episode.

The formalism used to model the environment and the agent's interactions within it in RL is the Markov Decision Process (MDP). An MDP is defined as a tuple \( (S, A, R, P, \gamma) \), where \( S \) is a finite set of states of the environment.
\( A \) is a finite set of actions that the agent can choose from.
\( P \) is the state transition probability matrix, with \( P(s'|s, a) \) representing the probability of transitioning from state \( s \) to state \( s' \) after the agent takes action \( a \).
\( R \) is a reward function, with \( R(s, a) \) representing the immediate reward the agent receives after taking action \( a \) in state \( s \).
\( \gamma \) is the discount factor, typically within the range \( 0 \leq \gamma \leq 1 \), which determines the present value of future rewards.

The agent's objective is to discover a policy \( \pi \), which is a mapping from states to the probabilities of selecting each possible action, \( \pi: S \to A \), that maximizes the expected sum of discounted rewards. The optimal policy \( \pi^* \) can be formally defined as:

\begin{equation}
\pi^* = \arg\max_{\pi} \mathbb{E}\left[\sum_{t=0}^{\infty} \gamma^t R(s_t, a_t) | s_0 = s, a_t = \pi(s_t)\right],
\end{equation}

RL algorithms, such as Q-learning\cite{r32}, double Q-learning \cite{r35} and proximal policy optimization, are employed to estimate \( \pi^* \) by leveraging the dynamics defined by the MDP.
\subsection{Soft Actor-Critic} 
The Soft Actor-Critic (SAC) algorithm \cite{r25} is an off-policy, actor-critic reinforcement learning algorithm that incorporates the principles of entropy maximization to balance exploration and exploitation. SAC employs two types of neural networks: soft Q-networks that approximate the soft Q-functions, and policy networks that generate probabilities distribution over actions. The policy network is trained to maximize the expected reward and entropy, leading to robust and efficient learning. The SAC algorithm optimizes the following entropy-augmented objective function:

\begin{equation}
J(\pi_\theta) = \mathbb{E}_{(s_t, a_t) \sim \rho_\pi} [r(s_t, a_t) + \alpha \mathcal{H}(\pi(\cdot|s_t))],
\end{equation}
where $\pi$ is the policy, $\theta$ are the parameters of the policy, $r(s_t, a_t)$ is the immediate reward for action $a_t$ in state $s_t$, $\rho_\pi$ is the distribution over states and actions under policy $\pi$, $\mathcal{H}$ is the policy entropy, and $\alpha$ is the entropy coefficient.

SAC uses two Q-networks, $Q_{\phi_1}$ and $Q_{\phi_2}$ to evaluate the policy. The objective of the Q-network, $J_Q(\phi_i)$, is defined as the expected squared error between the current Q-function and the target:

\begin{equation}
J_Q(\phi_i) = \mathbb{E}_{(s,a,r,s') \sim \mathcal{D}}\left[\left(Q_{\phi_i}(s,a) - y(r, s', \gamma)\right)^2\right],
\end{equation}

\begin{equation}
y(r, s', \gamma) = r + \gamma \left(\min_{i=1,2} Q_{\phi'_i}(s', \tilde{a}') - \alpha \log \pi_\theta(\tilde{a}'|s')\right),
\end{equation}
where $\tilde{a}'$ is action sampled from the current policy.

%The actor's objective, $J_\pi(\theta)$, is defined as the expected value of the Q-function under the current policy minus the scaled entropy of the policy:

%\begin{equation}
%J_\pi(\theta) = \mathbb{E}_{s \sim \mathcal{D}, a \sim \pi_\theta}\left[\min_{i=1,2}Q_{\phi_i}(s,a) - \alpha \log \pi_\theta(a|s)\right].
%\end{equation}

The policy update gradient is:

\begin{equation}
\nabla_\theta J(\pi_\theta) = \mathbb{E}_{s_t \sim \rho_\pi, a_t \sim \pi_\theta} [\nabla_\theta \log \pi_\theta(a_t|s_t) (Q_{\phi}(s_t, a_t) - \alpha \log \pi_\theta(a_t|s_t))],
\end{equation}

To stabilize learning, SAC employs soft target updates to slowly update the target network parameters $\phi'_i$:

\begin{equation}
\phi'_i \leftarrow \tau \phi_i + (1 - \tau) \phi'_i,
\end{equation}

where $\tau$ is a small number close to 0, indicating the rate at which the target network parameters are updated.

\subsection{Parameterized Soft Actor-Critic}

%In the SAC algorithm, actions are selected according to a stochastic policy. This policy is typically parameterized as a Gaussian distribution, allowing the model to capture a range of possible actions. The action at each timestep $t$ is sampled from this distribution, which is conditioned on the current state $s_t$:
%\begin{equation}
%a_t \sim \pi_\theta(\cdot|s_t) = \mathcal{N}(\mu_\theta(s_t), \sigma_\theta(s_t))
%\end{equation}
%where $\mu_\theta(s_t)$ and $\sigma_\theta(s_t)$ are the mean and standard deviation of the policy's Gaussian distribution, respectively, and are functions of the current state $s_t$ parameterized by $\theta$. This stochastic policy approach allows for exploration of the action space, which is an essential aspect of effective reinforcement learning.

%Building upon the conventional SAC algorithm, we introduce the Parameterized Action Space Soft Actor-Critic (PASAC) algorithm, designed to operate within environments that have both discrete and continuous action spaces. 
%In the PASAC algorithm, the output consists of a continuous action along with the probability of discrete actions. \(A_d = \{a_1, a_2, \ldots, a_k\}\), where each action \(a \in A_d\) is associated with a set of continuous parameters \(\mathbf{p}_a = \{p_{a1}, p_{a2}, \ldots, p_{ak}\} \subseteq \mathbb{R}^{k}\). The action space therefore includes continuous actions along with discrete actions, represented as \(A = \{a_{continue}, p_{a1}, p_{a2}, \ldots, p_{ak}\}\). 

In the SAC algorithm, actions are selected according to a stochastic policy. This policy is typically parameterized as a Gaussian distribution, allowing the model to capture a range of possible actions. Action at each timestep \(t\) is sampled from this distribution, which is conditioned on the current state \(s_t\):

\begin{equation}
a_t \sim \pi_\theta(\cdot|s_t) = \mathcal{N}(\mu_\theta(s_t), \Sigma_\theta(s_t)),
\end{equation}
where \(\mu_\theta(s_t)\) and \(\Sigma_\theta(s_t)\) are the mean and covariance of the policy's Gaussian distribution, respectively, and are functions of the current state \(s_t\) parameterized by \(\theta\). This stochastic policy approach allows exploration of the action space, which is an essential aspect of effective reinforcement learning.

Building upon the conventional SAC algorithm, we introduce the Parameterized Soft Actor-Critic (PASAC) algorithm, which is designed to operate within environments that have both discrete and continuous action spaces. In the PASAC algorithm, the policy's output consists of a continuous action along with the probability of discrete actions. Let \(A_d = \{a_1, a_2, \ldots, a_k\}\), where each action \(a_i \in A_d\) is associated with a set of continuous parameters \(\mathbf{p}_{a_i} = \{p_{a1}, p_{a2}, \ldots, p_{ak}\} \subseteq \mathbb{R}^k\). Therefore, the action space is represented as \(A = \{a_{continue}, a_1, a_2, \ldots, a_k\}\), including continuous and discrete actions.

\subsection{Parameterized Soft Actor-Critic with PID Lagrangian}
The Constrained Markov Decision Process (CMDP) \cite{r26} extends the MDP framework by augmenting it with constraints restricting the set of feasible policies. A CMDP is characterized by the expanded tuple \((S, A, R, P, \gamma, c, d)\), where \(c\) the cost function, and \(d\) the corresponding cost limit.

The objective of a CMDP is to optimize policy \(\pi\), yielding the highest expected sum of discounted rewards over trajectories, while keeping the expected sum of discounted costs within certain bounds. Formally, in a CMDP formulation, the RL problem finds the optimal policy \(\pi^*\) that solves:

\begin{equation}
\pi^* = \underset{\pi}{\text{arg max}}\ J_R(\pi) \quad \text{s.t. }  J_C(\pi) \leq d,
\end{equation}
where \( J_R(\pi) \) represents the expected reward for the policy \( \pi \), while \( J_C(\pi) \) denotes the cost associated with the policy \( \pi \). 

In this study, we address the constrained problem by employing the Lagrangian method, which allows us to convert a constrained problem into an unconstrained optimization problem. Lagrangian techniques are a well-established approach for tackling optimization problems that include constraints. Given a CMDP, the unconstrained problem can be written as:
\begin{equation}
\min_{\lambda \geq 0} \max_{\theta} L(\lambda, \theta) = \min_{\lambda \geq 0} \max_{\theta} \left[ J_R(\pi_\theta) - \lambda (J_C(\pi_\theta) - d) \right],
\end{equation}
where \( L \) is the Lagrangian and \( \lambda \geq 0 \) is the Lagrangian multiplier (a penalty coefficient).

In the traditional Lagrangian multiplier method, updates consider only integral control, which is related to the accumulation of constraint violations. Such updates can be conducted within the framework of the Lagrangian method by solving the dual problem, in which the multipliers are adjusted over time to satisfy the constraints.

The Lagrangian multiplier update formula (traditional integral control) can be represented as:
\begin{equation}
    \lambda_{k+1} =\text{max}(\lambda_{k} +  \alpha_{\lambda}(J_C(\pi_\theta) - d), 0 )
\end{equation}
where $\alpha_{\lambda} $ is the learning rate of $\lambda$

In the PID method, the dual update rule is enhanced by adding proportional (P) and derivative (D) controls to the existing integral (I) term, with the goal of reducing oscillations in system output and providing a quicker response to safety constraint violations. This PID Lagrangian approach is designed to overcome the limitations of previous methods by modifying the update rule to include P and D terms that are directly related to the constraint function. The new PID Lagrangian update rule is expressed as:

\begin{equation}
\lambda_{\text{new}} = \lambda_{\text{old}} +  \left( K_p e(t) + K_i \int e(t) \, dt + K_d \frac{d}{dt} e(t) \right)
\end{equation}
where \( e(t) =  J_C(\pi_\theta) - d \) is the constraint violation at time \( t \), with \( d \) being the target value for the constraint.
\( K_p \), \( K_i \), and \( K_d \) are the proportional, integral, and derivative gains, respectively.The proportional term \( K_p e(t) \) accounts for the current magnitude of the constraint violation, the integral term \( K_i \int e(t) \, dt \) considers the accumulated error over time, and the derivative term \( K_d \frac{d}{dt} e(t) \) takes into account the rate of change of the error. This combination helps to satisfy the constraints more quickly and smoothly during the learning process. The pseudocode of the PASAC-PIDLag algorithm is shown in Algorithm \ref{PASAC-PIDLag}.

\begin{algorithm}[H]
\caption{Parameterized Soft Actor-Critic with PID Lagrangian}\
\label{PASAC-PIDLag}
\begin{algorithmic}[1]
\State \textbf{Algorithm:}
\State Initialize $\theta, \phi_1, \phi_2,\phi'_1 \gets \phi_1, \phi'_2 \gets \phi_2, \mathcal{D}$ \Comment{Init parameters and replay buffer $\mathcal{D}$}
\State Initialize PID gains $K_p$, $K_i$, $K_d$, Lagrangian multiplier $\lambda$, target cost $d$
\State Initialize $J_{c,prev} \gets 0$, $I \gets 0$ \Comment{Init cost and integral term}
\For{each iteration}
    \For{each environment step}
        \State $a_{cont}, a_{disc} \sim \pi_\theta(\cdot|s)$ 
        \State $s', r, c \sim \text{Env}(s, a_{cont}, a_{disc})$ 
        \State $\mathcal{D} \gets \mathcal{D} \cup \{(s, a_{cont}, a_{disc}, r, c, s')\}$ \Comment{Store transition}
    \EndFor
    \For{each gradient step}
        \State $\{s, a_{cont}, a_{disc}, r, c, s'\} \sim \mathcal{D}$ \Comment{Sample batch}
        \State $\phi_i \gets \phi_i - \nabla_{\phi_i} J_Q(\phi_i)\quad \quad$for $i \in \{1,2\}$ \Comment{Update Q-function parameters}
        \State $\theta \gets \theta - \nabla_\theta J_\pi(\theta)$ \Comment{Update policy parameters}
        \State $\phi_i' \gets \tau\phi_i + (1 - \tau)\phi_i'\quad \quad$for $i \in \{1,2\}$ \Comment{Update target network parameters}
        \State $e \gets J_c - d$ 
        \State $I \gets I + e$ 
        \State $\Delta e \gets J_c - J_{c,prev}$ 
        \State $\lambda \gets \max(\lambda + K_p e + K_i I + K_d \Delta e, 0)$ \Comment{Update $\lambda$ using PID controller}
        \State $J_{c,prev} \gets J_c$ 
    \EndFor
\EndFor
\end{algorithmic}
\end{algorithm}

%%%%%%%%%%%%%%%%%%%%%%%%%%%%%%%%%%%%%%%%%%
\section{ Lane Change Problem Formulation}
\label{sec:application}

\subsection{Lane Change Environment}
\begin{figure}[H]
\includegraphics[width=1\linewidth]{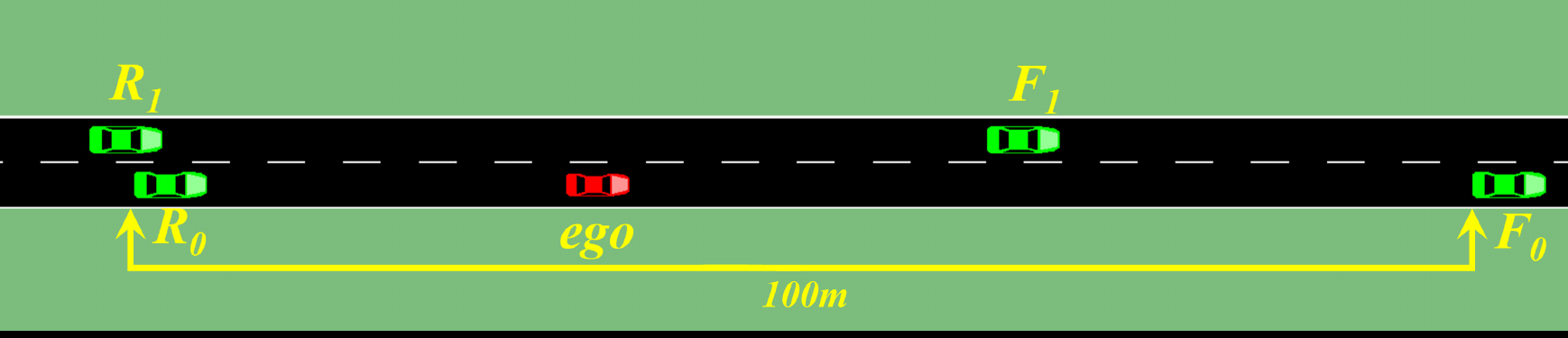}
\caption{Lane change environment created using SUMO, with the ego vehicle depicted in red and other traffic participants in green.\label{fig1}}
\end{figure}

The lane change environment is created in the Simulation of Urban Mobility (SUMO) \cite{r27} driving simulator. We use a two-lane road with a length of 1 km as our training road, and subsequently, testing is conducted on the same road. In this paper, the perception range of the vehicles is a circle with a radius of 200 m, and we assume that our vehicle can accurately perceive the status of all vehicles within its perception range, including itself. The other vehicles on the road have an initial speed of 8.33 m/s and a maximum speed of 16.67 m/s, and they use the IDM \cite{r28} model for longitudinal control, and the SL2015 \cite{r29} model for lateral control. In this study, we train with a traffic flow density of 15 \text{veh/km}. As shown in Figure \ref{fig1}, the red vehicle represents the ego vehicle, and the green vehicles represent other vehicles.
\unskip
\subsection{Environment State}
In this paper, the state is characterized by ten variables: the distance $d_{F_{0}}$ between the ego vehicle and the vehicle in front, the distance $d_{R_{0}}$ between the ego vehicle and the vehicle behind, the distance $d_{F_{1}}$ between the ego vehicle and the vehicle in front on the target lane, and the distance $d_{R_{1}}$ between the ego vehicle and the vehicle behind on the target lane, as well as the speeds $v_{F_{0}}$, $v_{F_{1}}$, $v_{R_{0}}$, $v_{R_{1}}$ of these four vehicles, and the speed $v_{\text{ego}}$ and acceleration $a_{\text{ego}}$ of the ego vehicle.

\begin{equation}
s = \{v_{F_1}, d_{F_1}, v_{R_1}, d_{R_1}, v_{F_0}, d_{F_0}, v_{R_0}, d_{R_0}, v_{ego}, a_{ego}\},
\end{equation}

\subsection{Control Action}
In this study, the continuous action of the control output is acceleration, and the discrete action is the lane-change decision. Vehicle dynamics and latency are not considered; hence, the vehicle instantaneously executes upon receiving an acceleration command or a lane-change decision. The update of the vehicle's velocity and position occurs with a time step of \(0.1\) seconds, while the lane-change decision is output every \(1\) second. Moreover, accounting for the actual vehicle's limits, the interval for continuous actions, is defined as \([a_{\min}, a_{\max}] = [-9.8, 5.0] \, \text{m/s}^2\), where \(a_{\min}\) and \(a_{\max}\) represent the minimum and maximum acceleration, respectively.

The action space is defined as a tuple \( A = (a_{\text{continuous}}, a_{\text{discrete}}) \), where:
\( a_{\text{continuous}} \) represents the continuous control of the vehicle's acceleration, bounded by \( a \in [-a_{\max}, a_{\max}] \).
\( a_{\text{discrete}} \) is the discrete lane-change decision, where \( a_{\text{discrete}} = 1 \) indicates a lane change to another lane, and \( a_{\text{discrete}} = 0 \) signifies maintaining the current lane.

\subsection{Reward and Cost}
In the context of autonomous vehicle control, reward and cost functions are designed to promote safe, efficient, and comfortable driving behaviors. These functions are itemized as follows:

\textbf{Reward:}
\begin{itemize}
  \item $r_{lc}$:The penalty $r_{lc}$ is for unnecessary lane change, which discourages erratic maneuvers.
  \item  $r_{spd}$: The reward $r_{spd}$ is for maintaining a target speed range and the penalty for deviations enforces speed limits.
  \item $r_{dis}$: The penalty $r_{dis}$ is for not maintaining safe following distances, which encourages the vehicle to maintain a safe distance from others.
  \item $r_{collision}$:The heavy penalty $r_{collision}$ is for collisions, which prioritizes safety.
  \item  $r_{jerk}$:The penalty $r_{jerk}$ is for sudden changes in acceleration, which ensures ride comfort.
\end{itemize}

\textbf{Cost:}
\begin{itemize}
  \item \textit{Time-To-Collision (TTC)}: Increases the cost when the potential for a collision is detected within a specified timeframe, promoting proactive collision avoidance.
\end{itemize}
The rewards and cost described above are included at each time step:

(1) This reward function aims to reduce meaningless lane change caused by the ego vehicle.
\begin{equation}
    r_{lc} =
    \begin{cases} 
    -4, & \text{if } d_{\text{front}} < 25 \, \text{meters} \text{ and lane change is decided} \\
    -20, & \text{if } d_{\text{front}} \geq 25 \, \text{meters} \text{ and lane change is decided}
    \end{cases}
\end{equation}

(2) To ensure that the ego vehicle maintains a speed within a specified range, and to facilitate its rapid approach to the destination when there are no vehicles ahead, we devised the following reward function. Here, \( d_{\text{safe}} \) represents the safe following distance from the vehicle ahead in the same lane, which is set to 25m in this study. \( v_{\text{limit}} \) denotes the minimum speed limit for the lane when the distance to the vehicle ahead exceeds the safe distance.
\begin{equation}
r_{\text{spd}} =
\begin{cases}
0.1 \times |v_{\text{ego}} - v_{\text{limit}}|,& v_{\text{ego}} \in [13.89 \text{ m/s}, 16.67 \text{ m/s}] \text{ and } d_f \geq d_{\text{safe}} \\
-0.1 \times |v_{\text{ego}} - v_{\text{limit}}|,& v_{\text{ego}} \notin [13.89 \text{ m/s}, 16.67 \text{ m/s}]\text{ and } d_f \geq d_{\text{safe}} 
\end{cases}
\end{equation}

(3) To facilitate the ego vehicle's acquisition of car-following behaviors and to mitigate the risk of collisions, we devised the following reward function predicated on the vehicle-to-vehicle distance metric:

\begin{equation}
r_{\text{dis}} = 
\begin{cases}
- 1 \cdot (d_{\text{safe}} - \min(d_{F0}, d_{R0})), & \text{if } d_{F0} \leq d_{\text{safe}} \text{ or } d_{R0} \leq d_{\text{safe}} \\
0, & \text{otherwise}
\end{cases}
\end{equation}
In the formula, \(d_{R0}\) represents the distance to the rear vehicle in the same lane, and \(d_{F0}\) denotes the distance to the forward vehicle in the same lane.

(4) To instruct the ego vehicle to autonomously navigate lane change while mitigating collision occurrences, a penalty of \(r_{\text{collision}} = -200\) is incurred following each collision event.

(5) To reduce the jerk during the ego vehicle's motion, we defined the following reward function: 

\begin{equation}
r_{\text{jerk}} = -0.005 \times \lvert a_{t} - a_{t-1} \rvert
\end{equation}
The term \(a_t\) represents the acceleration of the ego vehicle at the current time step, and \(a_{t-1}\) represents the acceleration of the ego vehicle at the previous time step.

(6) For safe reinforcement learning, we employ TTC as a cost metric.TTC is expressed as:

\begin{equation}
TTC = \frac{v_{ego} - v_{other}}{d_{relative}}
\end{equation}
Where \( v_{\text{ego}} \) represents the velocity of the ego vehicle, \( v_{\text{other}} \) denotes the velocity of other vehicles, and \( d_{\text{relative}} \) indicates the relative distance between the ego vehicle and other vehicles.When the TTC between the ego vehicle and either the leading or following vehicle is less than 2.7 seconds but greater than 0, the cost is incremented by 1; if the TTC is equal to or greater than 2.7 seconds or TTC is not calculable (due to no vehicle being present), the cost remains at 0. 

For the PASAC algorithm, the total reward at each timestep is given by:
\begin{equation}
r_{total} = r_{lc} + r_{spd} + r_{dis} + r_{jerk} + r_{collision}
\end{equation}

For the PASAC-PIDLag algorithm, the total reward and cost at each timestep is given by:
\begin{equation}
\begin{aligned}
r_{\text{total}} &= r_{\text{lc}} + r_{\text{spd}} + r_{\text{dis}} + r_{\text{jerk}} \\
\text{Cost} &= 
\begin{cases} 
\text{Cost} + 1 & \text{if } 0 < \text{TTC} < 2.7\,\text{s} \\
\text{Cost} & \text{if } \text{TTC} \geq 2.7\,\text{s or TTC is not calculable}
\end{cases}
\end{aligned}
\end{equation}

\section{Experiments and Results}
\label{sec:experiments}

In this section, we present the training results of the two algorithms under a traffic density of 15 \text{veh/km}, while also comparing the safety and optimality of the two algorithms under traffic densities of 10 \text{veh/km}, 15 \text{veh/km}, and 18 \text{veh/km}.

\subsection{Training}
Our training setup consisted of an NVIDIA RTX 3060 GPU and an Intel i7-12700F CPU, with each training session running for approximately 5 hours and covering 400,000 timesteps. The timestep interval was set at 0.1 seconds to better reflect real-world scenarios. Additionally, we initialized vehicles on the main road within a 50-meter buffer zone at the start of each episode.

\begin{figure}[H]
\begin{adjustwidth}{-\extralength}{0cm}
    \centering
    \includegraphics[width=1\linewidth]{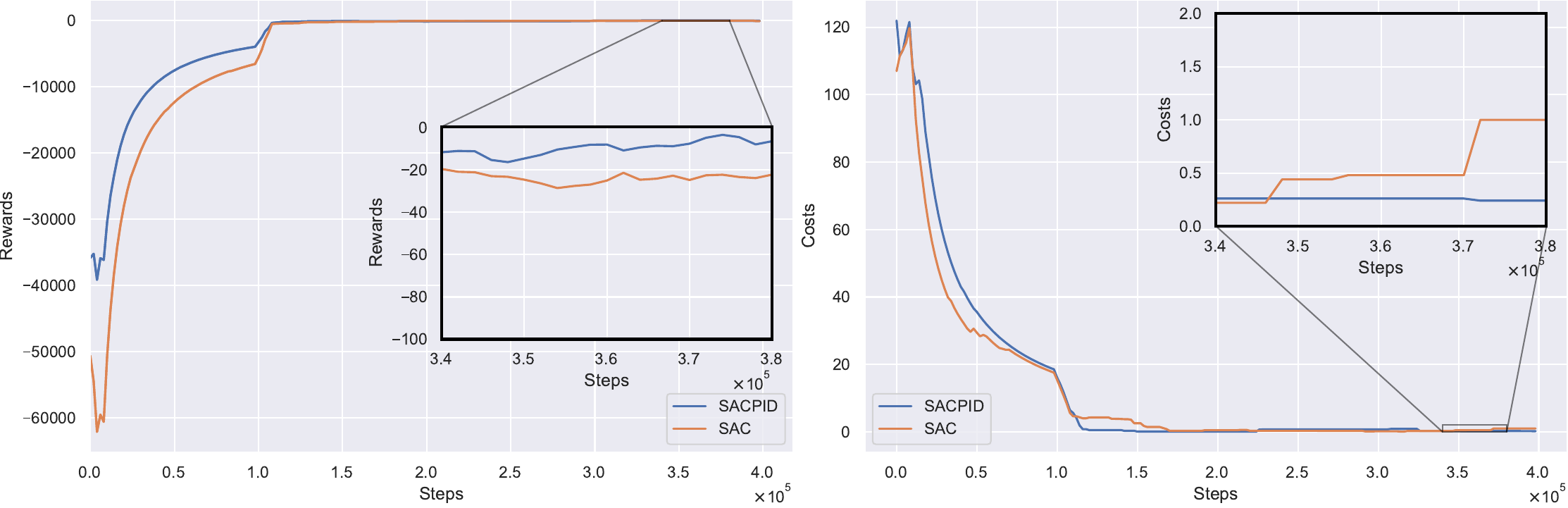}
\end{adjustwidth}
\caption{The training progress of the PASAC-PIDLag algorithm compared to the PASAC algorithm.\label{fig2}}
\end{figure}

The hyperparameter configurations for the PASAC-PIDLag and PASAC algorithms are listed in Table \ref{tab1}. Figure \ref{fig2} illustrates the training curves for both algorithms. It is important to note that, unlike PASAC-PIDLag, the PASAC algorithm does not incorporate cost in its gradient updates. From the training curves, we can discern that regardless of whether we consider reward or cost, the PASAC-PIDLag algorithm demonstrates superior performance.

\begin{table}[H]
\caption{The hyperparameter values for the PASAC-PIDLag and the PASAC algorithms.}
\begin{tabular}{p{6cm}p{3cm}p{3cm}}
\toprule
\textbf{Hyperparameter}	& \textbf{PASAC-PIDLag}	& \textbf{PASAC} \\
\midrule
Discount factor $\gamma$				& 0.99                & 0.99 \\
Temperature parameter $\alpha$				& 0.2     & 0.2 \\
The learning rate of Actor network					& 0.0001 & 0.0001 \\
The learning rate of Critic network					& 0.0003 & 0.0003 \\
Start learning steps				& 10000 & 10000 \\
The size of batch					& 256     & 256 \\
The size of replay buffer					& 1000000 & 1000000 \\
The soft update coefficient					& 0.005 & 0.005 \\
The $K_p$ of PID controller                    & 0.000002 &- \\
The $K_i$ of PID controller                    & 0.0000002 &- \\
The $K_d$ of PID controller                    & 0.0000001 &- \\
Tolerance of constraint violation                    & 0 &- \\
Initial value of lagrangian multiplier                    & 0.001 &- \\
%The exponential moving average alpha of the proportional term of the PID controller &0.95\\
%The exponential moving average alpha of the derivative term of the PID controller &0.95\\
\bottomrule
\end{tabular}
\label{tab1}
%\noindent{\footnotesize{\textsuperscript{1} Tables may have a footer.}}
\end{table}

\subsection{Testing}
In our experiments, we evaluated the performance of the trained policy over 400 episodes under a traffic density of 15 \text{veh/km}, encompassing approximately 300,000 timesteps. At the onset of each episode, the initial velocity of the ego vehicle was set to 8.33 m/s (equivalent to 30 km/h). Moreover, to assess the generalizability of our approach, we also conducted tests on the aforementioned strategy at traffic densities of 10 \text{veh/km} and 18 \text{veh/km}.

\subsection{Comparison and Analysis}

Based on the results obtained from the dataset of 400 test episodes, as shown in Table \ref{tab2}, it is evident that the PASAC-PIDLag algorithm outperforms the PASAC algorithm on multiple evaluation metrics. The PASAC-PIDLag algorithm exhibits a notably lower collision rate, indicating a safer driving policy adept at mitigating the risk of accidents more effectively. In addition, this algorithm necessitates fewer lane-change maneuvers, suggesting a more stable and efficient driving behavior with the potential to diminish disruptive actions within the traffic flow.
In terms of velocity, the PASAC-PIDLag algorithm achieves a higher average speed, which is a pivotal factor in enhancing the rate of transport. Moreover, the metric of jerk is significantly reduced for the PASAC-PIDLag algorithm. A lower jerk signifies a smoother driving experience, which translates to increased comfort for the occupants.
Upon comprehensive consideration of these performance indicators, the PASAC-PIDLag algorithm surpasses the PASAC algorithm in terms of both optimality and safety.

\begin{table}[H] 
\caption{The results under a traffic flow density of 15 \text{veh/km}.}
\newcolumntype{A}{>{\centering\arraybackslash}X}
\begin{tabularx}{\textwidth}{CCC}
\toprule
400 Episodes & \textbf{PASAC-PIDLag} & \textbf{PASAC} \\
\midrule
\textbf{Average reward}  & \textbf{0.0053} & \textbf{-0.1024} \\

\textbf{Collision rate}  & \textbf{0\%} & \textbf{1\%} \\

Average acceleration (m/s\textsuperscript{2}) & 0.078 & 0.073 \\

Average speed (m/s) & 14.36 & 14.04 \\

Average jerk (m/s\textsuperscript{3}) & 0.315 & 0.415 \\

Lane change times & 137 & 146 \\

\bottomrule
\end{tabularx}
%\noindent{\footnotesize{\textsuperscript{1} Tables may have a footer.}}
\label{tab2}
\end{table}

\begin{figure}[H]
\begin{adjustwidth}{-\extralength}{0cm}
    \centering
    \includegraphics[width=6cm]{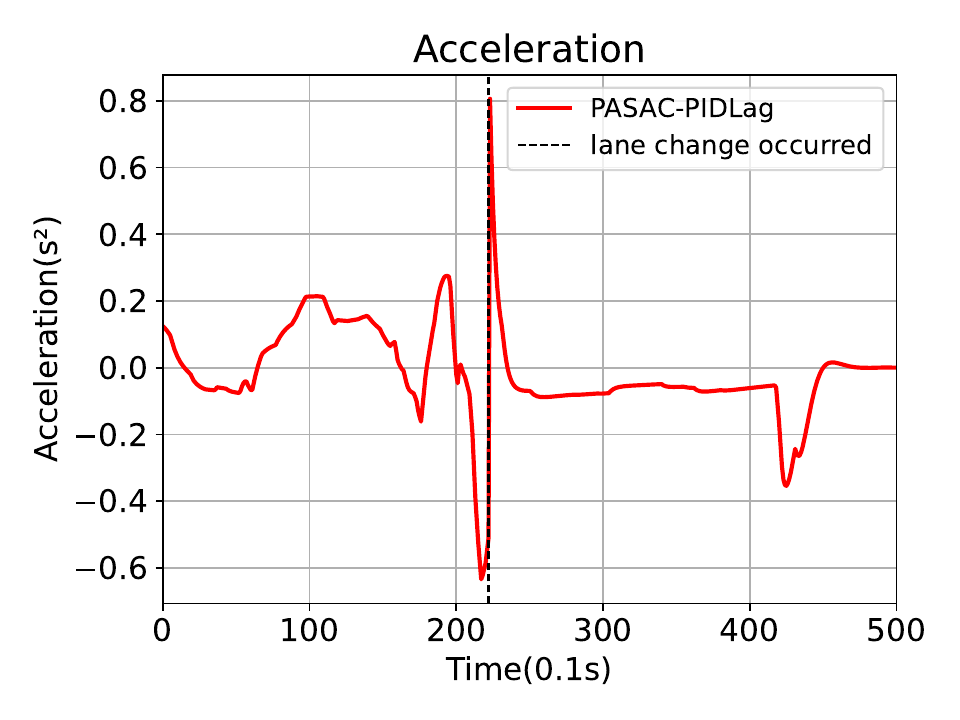}
    \includegraphics[width=6cm]{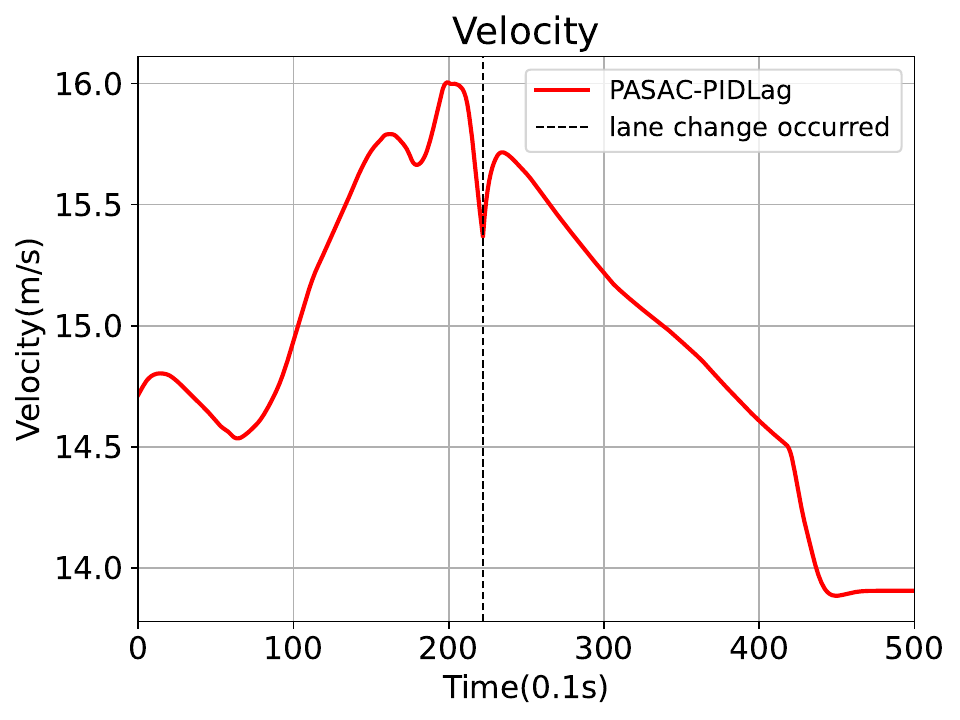}
    \includegraphics[width=6cm]{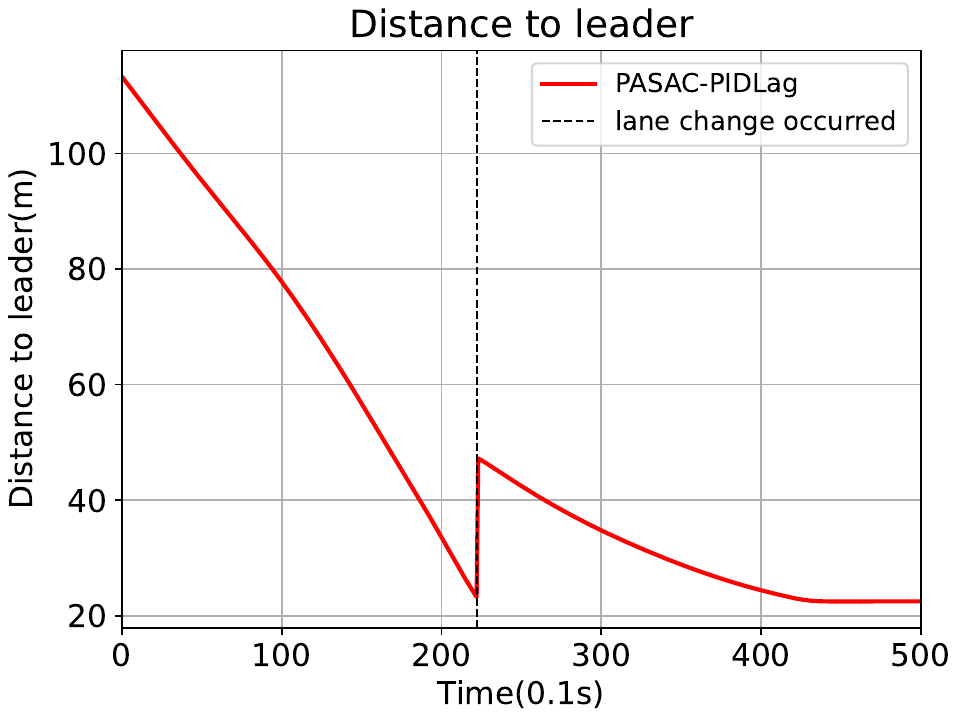}
\end{adjustwidth}
\caption{The vehicle's velocity, acceleration, and distance of the leader vehicle under the regulation of PASAC-PIDLag algorithmic controls. A black dashed line traverses the graphs, symbolizing the execution of a successful lane change by the ego vehicle.\label{fig3}}
\end{figure}

\begin{figure}[H]
%\begin{adjustwidth}{-\extralength}{0cm}
    \centering
    \includegraphics[width=1.0\linewidth]{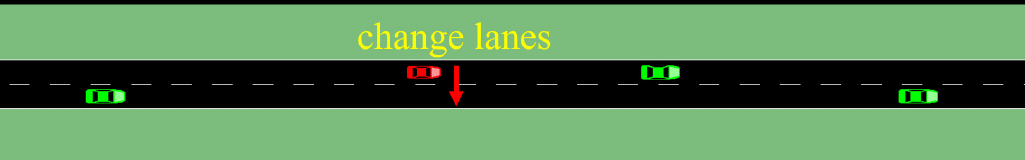}
%\end{adjustwidth}    
\caption{The figure illustrates a successful lane change maneuver executed by a vehicle under the control of the PASAC-PIDLag algorithm, where the red vehicle is denoted as the ego car, and the green vehicles represent the surrounding traffic.\label{fig4}}
\end{figure}

Figure \ref{fig3} provides an analysis of a lane-changing episode under the PASAC-PIDLag algorithm. Subsequent to this lane-change event, there is an immediate and discernible change in the distance to the preceding vehicle, indicative of the completion of the lane change.  The graph detailing relative distance demonstrates that the vehicle initiates the lane change maneuver when it is at a safe following distance of approximately 25 meters. Moreover, the velocity graph depicts a modest escalation in the ego vehicle's speed following the lane change, which is shortly followed by a decrease. Figure \ref{fig4} presents an example of a successful lane change maneuver executed by the PASAC-PIDLag algorithm.

\begin{figure}[H]
\begin{adjustwidth}{-\extralength}{0cm}
\centering
\includegraphics[width=6cm]{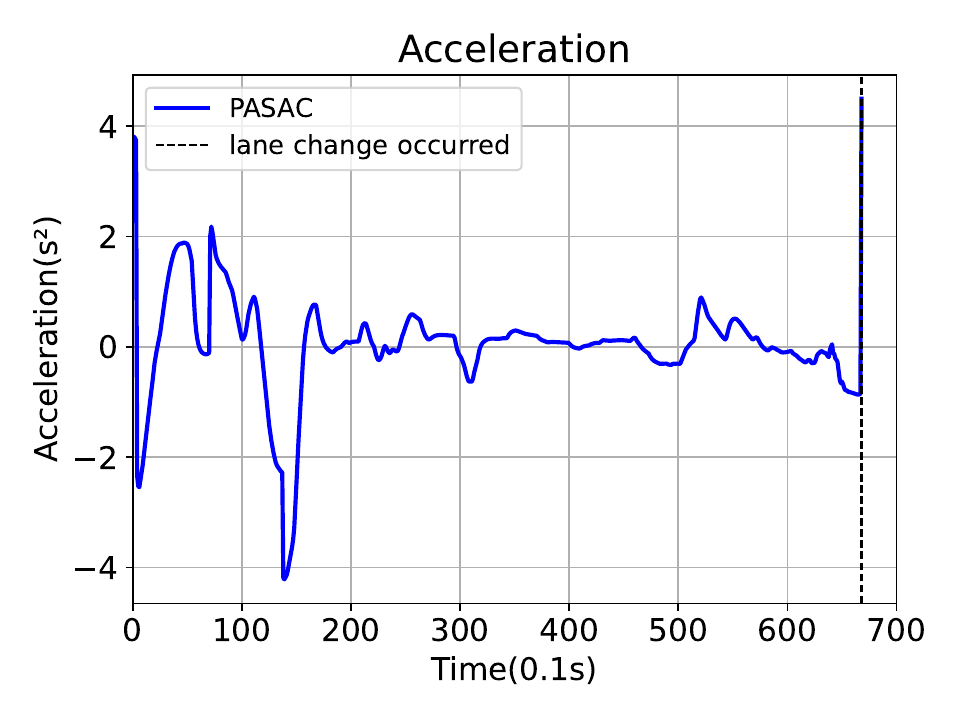}
\includegraphics[width=6cm]{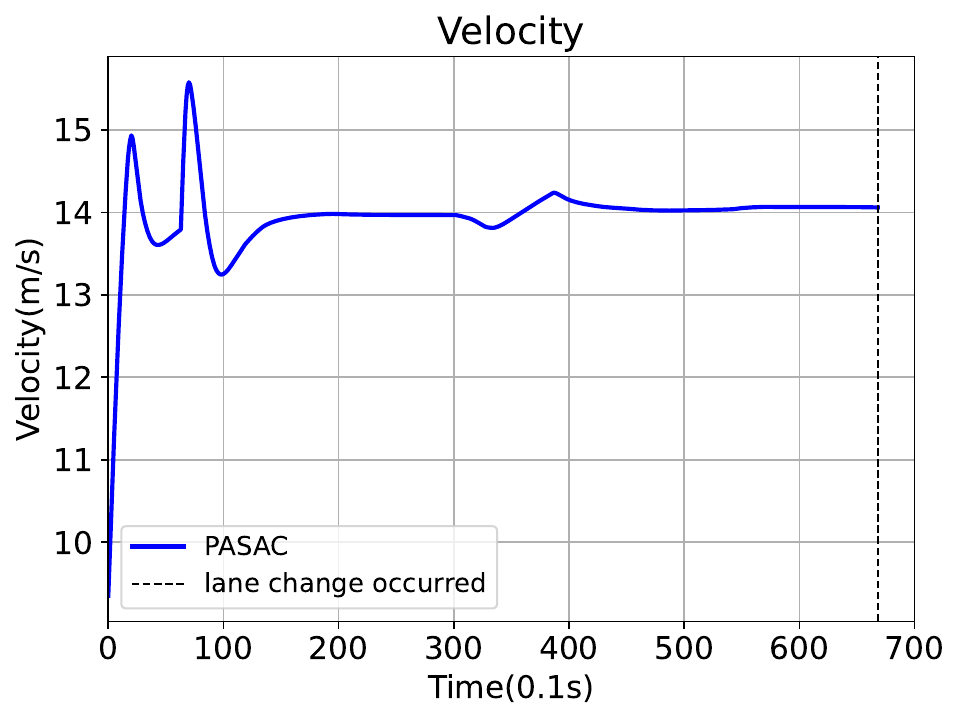}
\includegraphics[width=6cm]{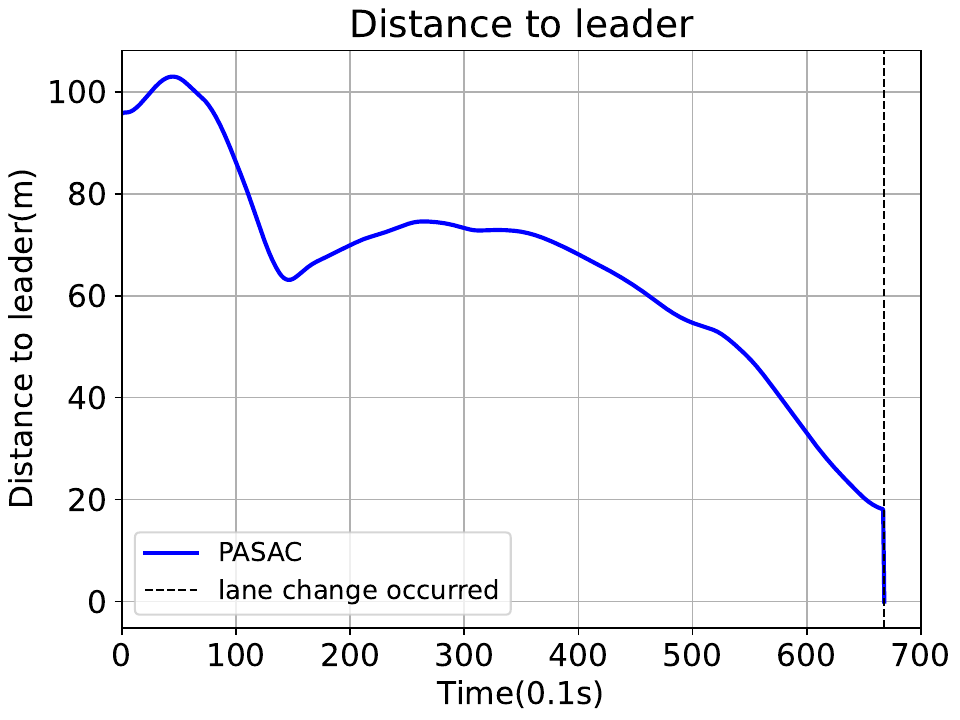}
\end{adjustwidth}
\caption{The velocity, acceleration, and lead vehicle distance during a collision event due to lane changing under the PASAC algorithm.\label{fig5}}
\end{figure}

\begin{figure}[H]
    \centering
    \includegraphics[width=1\linewidth]{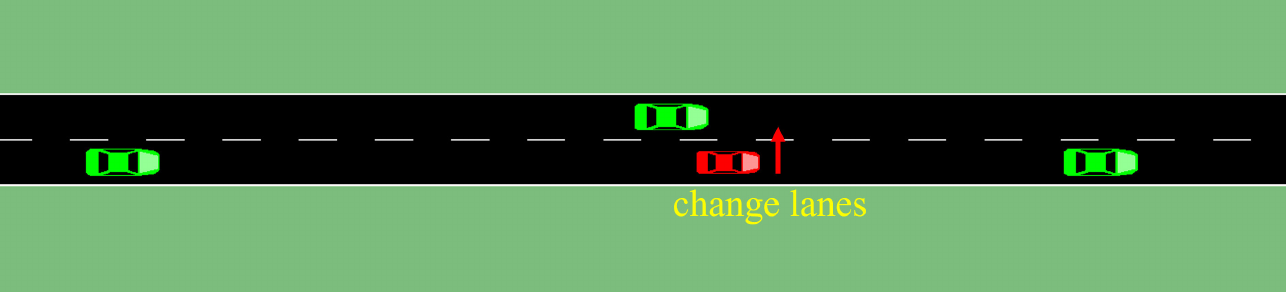}
    \caption{A collision incident during a lane change maneuver controlled by the PASAC algorithm, where the red vehicle represents the ego car and the green vehicle represents other traffic participant.}
    \label{fig6}
\end{figure}

Figure \ref{fig5} depicts an episode of collision occurrence within the PASAC algorithm framework, in which the ego vehicle collides after executing a lane change. The data presented in the figure reveal that the ego vehicle was steadily closing in on the vehicle ahead until the following distance diminished to 19m, which triggered a decision to change lanes. At this juncture, the presence of another vehicle on the target lane led to a collision. Figure \ref{fig6} displays an instance of a lane change maneuver resulting in a collision, as directed by the PASAC algorithm.


\begin{figure}[H]
\begin{adjustwidth}{-\extralength}{0cm}
\centering
\includegraphics[width=6cm]{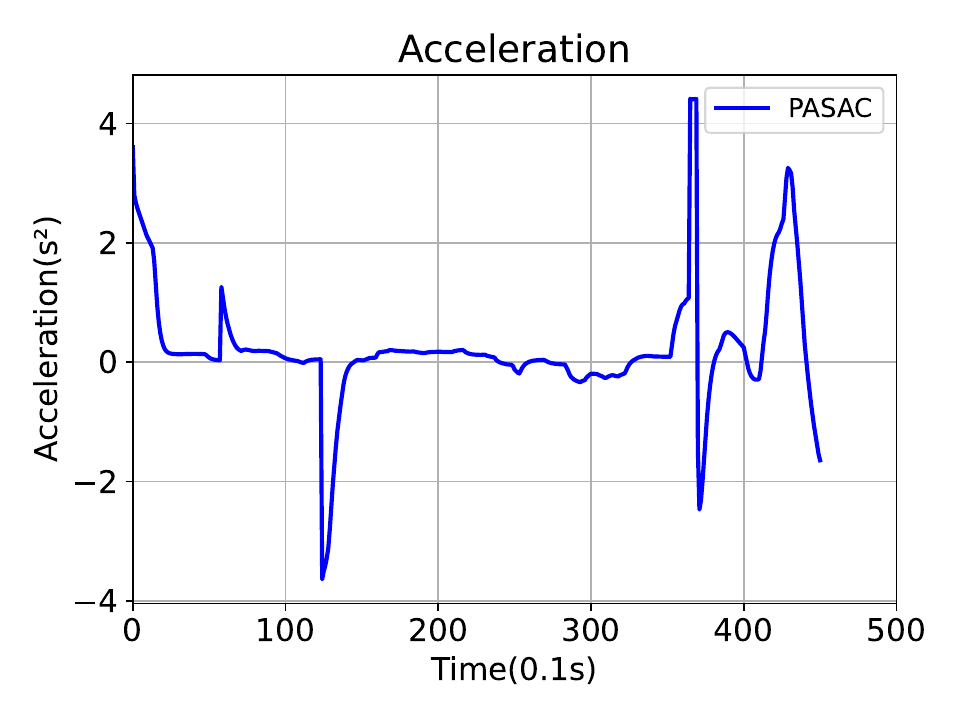}
\includegraphics[width=6cm]{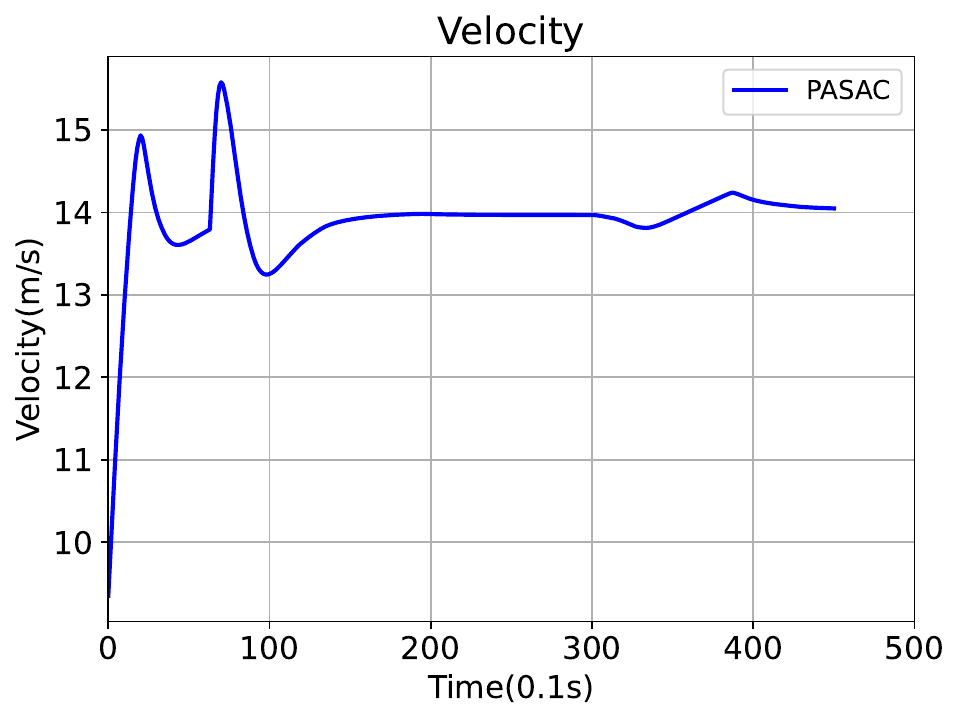}
\includegraphics[width=6cm]{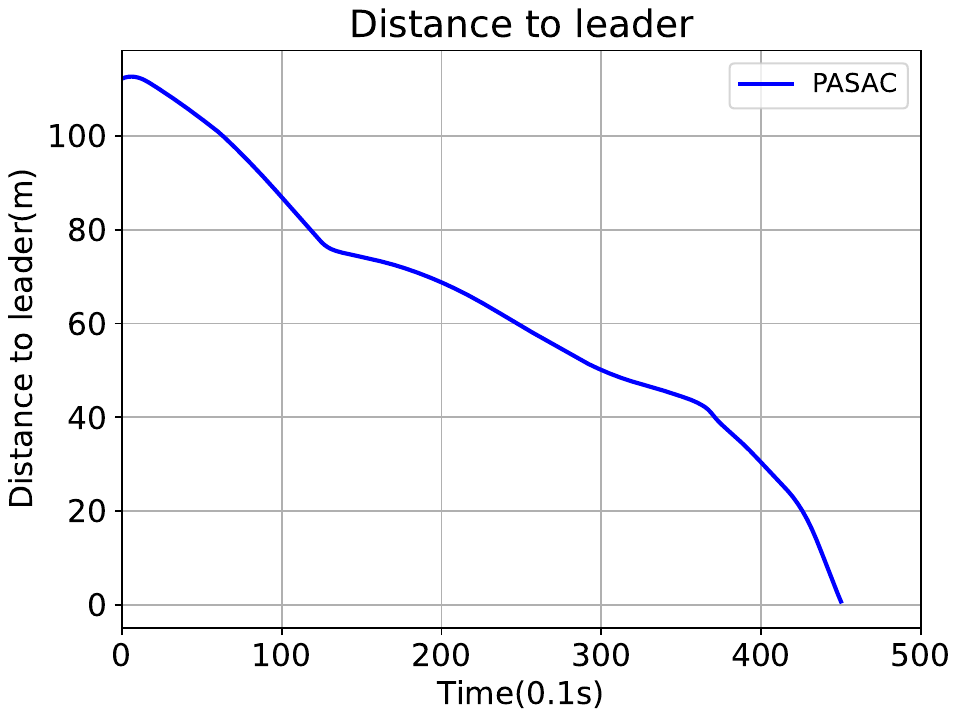}
\end{adjustwidth}
\caption{The velocity, acceleration, and lead vehicle distance during a collision event due to accelerating under the PASAC algorithm.\label{fig7}}
\end{figure}

\begin{figure}[H]
    \centering
    \includegraphics[width=1\linewidth]{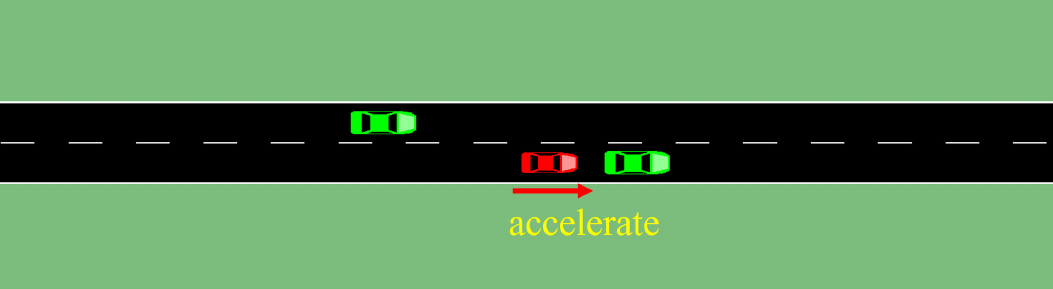}
    \caption{A collision incident during a car-following scenario controlled by the PASAC algorithm, where the red vehicle represents the ego car, and the green vehicle represents another traffic participant.}
    \label{fig8}
\end{figure}

Figure \ref{fig7} illustrates another scenario in which a collision occurs under the PASAC algorithm, where the ego vehicle collides during the car-following process. The data and the figure reveal that the ego vehicle approaches the lead vehicle at a constant speed; however, due to the presence of a vehicle on the adjacent lane, the ego vehicle is unable to change lanes, resulting in a collision. Figure \ref{fig8} presents an example of a collision involving an ego vehicle trained using the PASAC algorithm in a car-following scenario. A comparison of lane-changing decision between PASAC and PASAC-PIDLag demonstrates that the strategy derived from the PASAC algorithm is sometimes incapable of effectively balancing the decision related to lane changing and car following under certain conditions.

\subsection{Generalization Analysis}

To evaluate the generalizability of the proposed algorithm, we conducted tests under a traffic density of 10 \text{veh/km}, and the results are presented in Table \ref{tab3}. The data presented in Table \ref{tab3} reveal that at such a reduced vehicular density, both algorithms demonstrated the ability to maintain a collision rate of zero. Notwithstanding this equivalence in safety, the PASAC-PIDLag algorithm surpassed its counterpart PASAC by securing a greater average reward, attaining a higher mean velocity, and exhibiting a lower jerk metric. These findings imply that PASAC-PIDLag not only meets safety benchmarks but also excels in performance, offering an enhanced level of optimality over the PASAC algorithm.
\begin{table}[H] 
\caption{The generalization results under a traffic flow density of 10 \text{veh/km}.}
\newcolumntype{B}{>{\centering\arraybackslash}X}
\begin{tabularx}{\textwidth}{CCC}
\toprule
\textbf{400 Episodes} & \textbf{PASAC-PIDLag} & \textbf{PASAC} \\
\midrule
\textbf{Average reward}  & \textbf{0.0546} & \textbf{-0.0226} \\

\textbf{Collision rate} & \textbf{0\%} & \textbf{0\%} \\
Average acceleration (m/s\textsuperscript{2}) & 0.082 & 0.076 \\

Average speed (m/s) & 14.51 & 14.06 \\

Average jerk (m/s\textsuperscript{3}) & 0.290 & 0.329 \\

Lane change times & 82 & 78 \\

\bottomrule
\end{tabularx}
%\noindent{\footnotesize{\textsuperscript{1} Tables may have a footer.}}
\label{tab3}
\end{table}

Our final series of tests were conducted at a traffic flow density of 18 \text{veh/km}, the results outlined in Table \ref{tab4} reveal that at this higher density, the collision rate of the PASAC-PIDLag algorithm remained lower than that of PASAC. Furthermore, the PASAC-PIDLag algorithm demonstrated superiority in all measured metrics, including average reward, average speed, and average jerk.

\begin{table}[H] 
\caption{The generalization results under a traffic flow density of 18 \text{veh/km}.}
\newcolumntype{D}{>{\centering\arraybackslash}X}
\begin{tabularx}{\textwidth}{CCC}
\toprule
\textbf{400 Episodes} & \textbf{PASAC-PIDLag} & \textbf{PASAC} \\
\midrule
\textbf{Average reward}  & \textbf{-0.856} & \textbf{-2.999} \\

\textbf{Collision rate}  & \textbf{0.75\%} & \textbf{2.5\%} \\

Average acceleration (m/s\textsuperscript{2}) & 0.07 & 0.068 \\

Average speed (m/s) & 14.17 & 13.98 \\

Average jerk (m/s\textsuperscript{3}) & 0.290 & 0.329 \\

Lane change times & 226 & 164 \\

\bottomrule
\end{tabularx}
%\noindent{\footnotesize{\textsuperscript{1} Tables may have a footer.}}
\label{tab4}
\end{table}
    
\section{Conclusion}
\label{sec:conclusions}

In this paper, we introduce PASAC-PIDLag, a hybrid action space safe reinforcement learning algorithm, specifically applied to the scenario of autonomous lane-change. This method represents a novel approach that aims to enhance both safety and optimality in the application of reinforcement learning in the autonomous driving domain. We also compare it with its unsafe version PASAC. Both algorithms were trained and tested under traffic flow density scenarios of 0.15\text{veh/km} and underwent generalization tests at densities of 0.10\text{veh/km} and 0.18\text{veh/km}. The results indicate that in the absence of modeling errors, at a density of 15 \text{veh/km}, the strategy trained by the PASAC-PIDLag algorithm managed to maintain zero collisions. However, the collision rate for the PASAC algorithm was observed to be 1\%. The PASAC algorithm was observed to encounter two types of collisions at a density of 15 \text{veh/km}. Because the reward structure in this study involves both lane changing and car following, it inherently presents a multi-objective optimization problem, which may lead to collisions arising from unsuccessful lane changing or car following maneuvers.
Both algorithms achieved zero collisions at a traffic flow density of 10 \text{veh/km}. At a higher density of 18 \text{veh/km}, the collision rate of the PASAC-PIDLag algorithm was lower than that of the PASAC algorithm. Across the three traffic densities, the PASAC-PIDLag algorithm consistently achieved higher average speeds, lower average jerk, and greater average rewards. Overall, the PASAC-PIDLag algorithm shows superior performance with respect to safety and optimality. In future work, we aim to advance the application of safe reinforcement learning in actual vehicles by conducting hardware-in-the-loop simulation experiments.
% Example of a figure that spans the whole page width. The same concept works for tables, too.

%\begin{listing}[H]
%\caption{Title of the listing}
%\rule{\columnwidth}{1pt}
%\raggedright Text of the listing. In font size footnotesize, small, or normalsize. Preferred format: left aligned and single spaced. Preferred border format: top border line and bottom border line.
%\rule{\columnwidth}{1pt}
%\end{listing}

%%%%%%%%%%%%%%%%%%%%%%%%%%%%%%%%%%%%%%%%%%
\authorcontributions{Conceptualization, Y.L.; methodology, R.X.,and J.X.;formal analysis,R.X. and Y.L.; investigation,R.X.,Y.L.; data curation, R.X., L.X.; writing—original draft preparation, R.X.; writing—review and editing, Y.L., R.X.and L.X.; supervision, Y.L.; All authors have read and agreed to the published version of the manuscript}

\funding{ Not applicable}

\institutionalreview{ Not applicable}

\informedconsent{ Not applicable}

\dataavailability{The data can be obtained upon reasonable request from the first author} 

% Only for journal Nursing Reports
%\publicinvolvement{Please describe how the public (patients, consumers, carers) were involved in the research. Consider reporting against the GRIPP2 (Guidance for Reporting Involvement of Patients and the Public) checklist. If the public were not involved in any aspect of the research add: ``No public involvement in any aspect of this research''.}

% Only for journal Nursing Reports
%\guidelinesstandards{Please add a statement indicating which reporting guideline was used when drafting the report. For example, ``This manuscript was drafted against the XXX (the full name of reporting guidelines and citation) for XXX (type of research) research''. A complete list of reporting guidelines can be accessed via the equator network: \url{https://www.equator-network.org/}.}

%\acknowledgments{In this section you can acknowledge any support given which is not covered by the author contribution or funding sections. This may include administrative and technical support, or donations in kind (e.g., materials used for experiments).}

\conflictsofinterest{The authors declare that they have no known competing financial interest or personal relationships that could have appeared to influence the work reported in this paper.} 

%%%%%%%%%%%%%%%%%%%%%%%%%%%%%%%%%%%%%%%%%%

%%%%%%%%%%%%%%%%%%%%%%%%%%%%%%%%%%%%%%%%%%
\begin{adjustwidth}{-\extralength}{0cm}

\reftitle{References}


\bibliography{ref.bib}

%=====================================
% References, variant B: internal bibliography
%=====================================

% If authors have biography, please use the format below
%\section*{Short Biography of Authors}
%\bio
%{\raisebox{-0.35cm}{\includegraphics[width=3.5cm,height=5.3cm,clip,keepaspectratio]{Definitions/author1.pdf}}}
%{\textbf{Firstname Lastname} Biography of first author}
%
%\bio
%{\raisebox{-0.35cm}{\includegraphics[width=3.5cm,height=5.3cm,clip,keepaspectratio]{Definitions/author2.jpg}}}
%{\textbf{Firstname Lastname} Biography of second author}

% For the MDPI journals use author-date citation, please follow the formatting guidelines on http://www.mdpi.com/authors/references
% To cite two works by the same author: \citeauthor{ref-journal-1a} (\citeyear{ref-journal-1a}, \citeyear{ref-journal-1b}). This produces: Whittaker (1967, 1975)
% To cite two works by the same author with specific pages: \citeauthor{ref-journal-3a} (\citeyear{ref-journal-3a}, p. 328; \citeyear{ref-journal-3b}, p.475). This produces: Wong (1999, p. 328; 2000, p. 475)

%%%%%%%%%%%%%%%%%%%%%%%%%%%%%%%%%%%%%%%%%%
%% for journal Sci
%\reviewreports{\\
%Reviewer 1 comments and authors’ response\\
%Reviewer 2 comments and authors’ response\\
%Reviewer 3 comments and authors’ response
%}
%%%%%%%%%%%%%%%%%%%%%%%%%%%%%%%%%%%%%%%%%%
\PublishersNote{}
\end{adjustwidth}
\end{document}